\documentclass[runningheads]{llncs}
\usepackage[T1]{fontenc}
\usepackage{graphicx}
\usepackage{algorithm}
\usepackage{algorithmic}
\usepackage{amsfonts,amsmath}
\usepackage{pifont}
\usepackage{amssymb}     
\usepackage{xcolor}
\newcommand{\xmark}{\ding{55}}
\usepackage[symbol]{footmisc}
\newcommand\blfootnote[1]{%
  \begingroup
  \renewcommand\thefootnote{}\footnote{#1}%
  \addtocounter{footnote}{-1}%
  \endgroup
}
\begin{document}
\title{SemFaceEdit: Semantic Face Editing on Generative Radiance Manifolds}
%
%
\author{Shashikant Verma\orcidID{0000-0002-9862-1379} \and
Shanmuganathan Raman$^*$\orcidID{0000-0003-2718-7891}}
\authorrunning{S. Verma et al.}
%
\institute{CVIG Lab, Indian Institute of Technology Gandhinagar, India\\
\email{\{shashikant.verma, shanmuga\}@iitgn.ac.in}\\
}

\maketitle              
\begin{abstract}
Despite multiple view consistency offered by 3D-aware GAN techniques, the resulting images often lack the capacity for localized editing. In response, generative radiance manifolds emerge as an efficient approach for constrained point sampling within volumes, effectively reducing computational demands and enabling the learning of fine details. This work introduces SemFaceEdit, a novel method that streamlines the appearance and geometric editing process by generating semantic fields on generative radiance manifolds. Utilizing latent codes, our method effectively disentangles the geometry and appearance associated with different facial semantics within the generated image. In contrast to existing methods that can change the appearance of the entire radiance field,  our method enables the precise editing of particular facial semantics while preserving the integrity of other regions. Our network comprises two key modules: the Geometry module, which generates semantic radiance and occupancy fields, and the Appearance module, which is responsible for predicting RGB radiance. We jointly train both modules in adversarial settings to learn semantic-aware geometry and appearance descriptors. The appearance descriptors are then conditioned on their respective semantic latent codes by the Appearance Module, facilitating disentanglement and enhanced control. Our experiments highlight SemFaceEdit's superior performance in semantic field-based editing, particularly in achieving improved radiance field disentanglement. 

\end{abstract}
%
%
%


\section{Introduction}
\blfootnote{$^*$ This work is supported by Jibaben Patel Chair in AI.}
In recent years, there has been significant interest in vision and graphics in generating visually captivating and photo-realistic images. 
Notably, 3D aware Generative Adversarial Networks incorporating adversarial learning \cite{goodfellow2014generative} and Neural Radiance Fields \cite{mildenhall2021nerf} have made remarkable progress in producing multiple-view images that closely resemble real photographs.
Despite advancements, 3D aware Generative Adversarial Networks (GANs) \cite{or2022stylesdf,chan2021pi,gu2021stylenerf} still struggle to match the resolution of state-of-the-art 2D GAN models. 
This limitation arises from the high computational demands of learning volumetric representations. Recently proposed methods \cite{chan2022efficient,deng2022gram} try to efficiently learn radiance fields by a tri-plane representation and learnable radiance manifolds, respectively.
Nevertheless, these methods lack control over learned geometry, thus posing challenges in decoupling appearance from geometry.

To decouple facial geometric and appearance attributes, FENeRF \cite{sun2022fenerf} adopts $\pi$-GAN \cite{chan2021pi} and learns facial semantic volume aligned with geometry. The learned facial semantics act as an intermediary, offering control over geometry and appearance. 
However, implementing local editing with FENeRF necessitates time-consuming and resource-intensive retraining. 
Addressing this limitation, \cite{jiang2022nerffaceediting} proposes an AdaIN-based Controllable Appearance Module (CAM) and a geometry decoder.
This approach allows for independent control of appearance from geometry. 
Unlike FENeRF, this approach requires training a geometry decoder on learned tri-plane representations from \cite{chan2022efficient} to obtain the semantic volume for editing. 
Consequently, the CAM module in \cite{jiang2022nerffaceediting} grants control over the appearance of the entire generated portrait.
While introducing the geometry decoder branch facilitates disentangled face editing, it does not provide semantic control over appearance.

In this work, we introduce SemFaceEdit, building upon efficient Generative Radiance Manifolds (GRAM) \cite{deng2022gram}. GRAM employs 2D manifolds to constrain point sampling and radiance field learning, jointly trained with GAN. 
This constrained approach reduces the computational workload and enables fine detail learning by confining point sampling and radiance learning within a reduced space. 
In contrast to GRAM, our approach learns both semantically disentangled geometry and appearance in a unified manner. 
We introduce two modules: the Geometry Module and the Appearance Module. 
The Geometry Module predicts the semantic radiance field within a volume, which the Appearance Module subsequently uses to condition the appearance semantically. 
Unlike existing methods that can entirely change the appearance or geometry of a radiance field, our proposed approach enables semantic-specific editing while preserving the integrity of other semantic information in terms of geometry or appearance. 
Our method facilitates changes in geometry and appearance within a single photograph and enables seamless transfer of hairstyle or other semantic facial attributes from one source reference to another.

The main contributions of this work are summarized as follows.
\begin{itemize}
    \item We present a novel framework based on generative radiance manifolds, allowing for semantic control over the geometry and appearance by manipulating latent spaces specific to each semantic attribute.
    \item We design a differentiable Semantic Volume masking layer, which learns to segregate each point in radiance volume into semantic groups. This layer enables targeted appearance transfer concerning aspects like colours and hues and selective geometry transfer for various semantic regions.
    \item Our method excels in geometry and appearance control, allowing for the seamless transfer of geometry or appearance associated with one semantic label in a radiance field to another.
\end{itemize}

\begin{figure}[t]
\begin{center}
    \includegraphics[width=\linewidth]{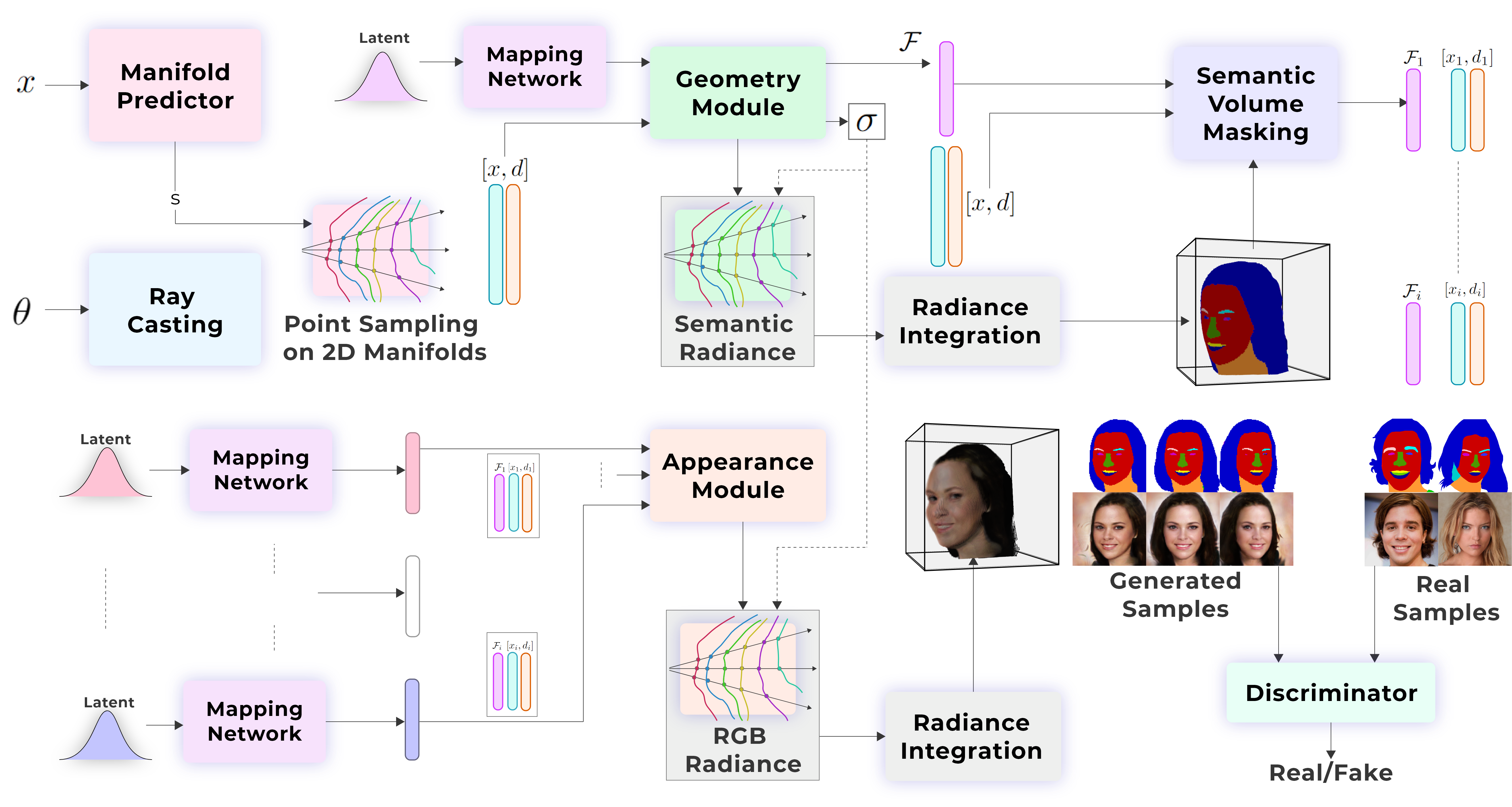}
\end{center}
    \caption{
    An overview of our proposed framework. 
    We sample points in volume by determining intersections of casted rays with isosurfaces predicted by Manifold Predictor \cite{deng2022gram}. 
    Subsequently, the Geometry Module conditions these points using a latent vector sampled from a Gaussian distribution, resulting in diverse predictions for occupancy ($\sigma$), semantic radiance, and Appearance Descriptor $\mathcal{F}$. 
    To segregate points and their appearance descriptors based on semantic classes, we employ the Semantic Volume Masking layer, which relies on semantic radiance information. 
    The Appearance Module then utilizes different latent codes to condition each set of points and predicts RGB Radiance. Selectively conditioning Appearance Descriptors $\mathcal{F}$ based on semantics enables the Appearance Module to independently modify a specific semantic appearance.
    }
    \label{fig:arch}
\end{figure}

\begin{figure}[t]
    \begin{center}
        \includegraphics[width=\linewidth]{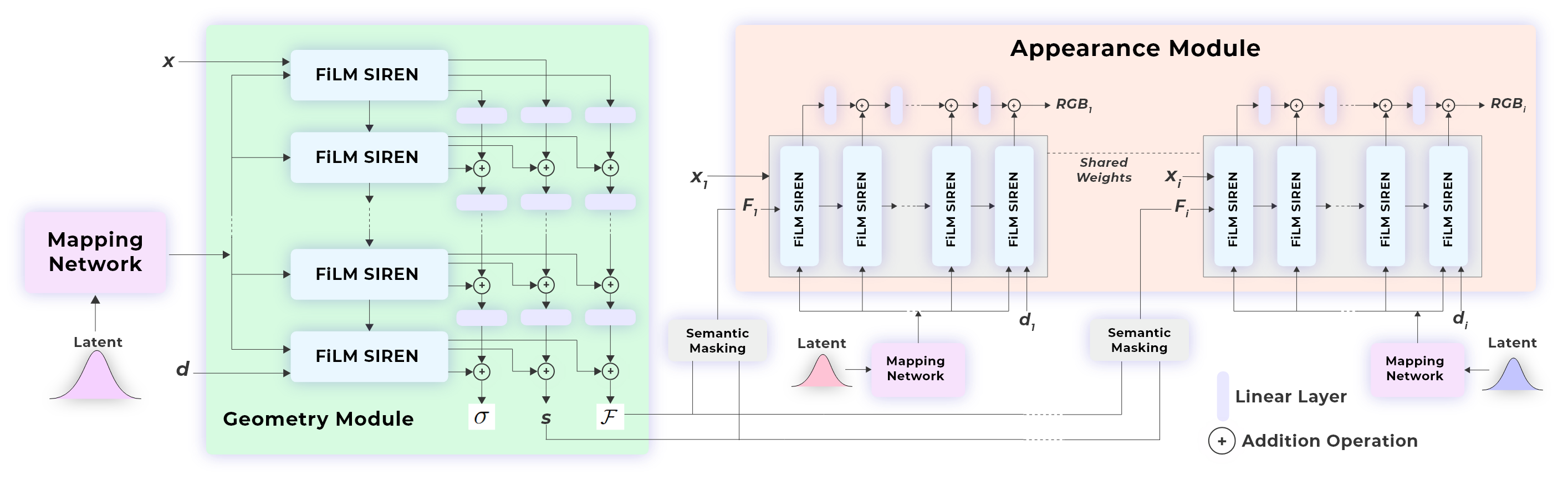}
    \end{center}
    \vspace{-0.4 cm}
    \caption{The network architecture of Geometry Module and Appearance Module.}
    \label{fig:archga}
    \vspace{-0.2 cm}
\end{figure}
\begin{figure}[t]
\begin{center}
    \includegraphics[width=\linewidth]{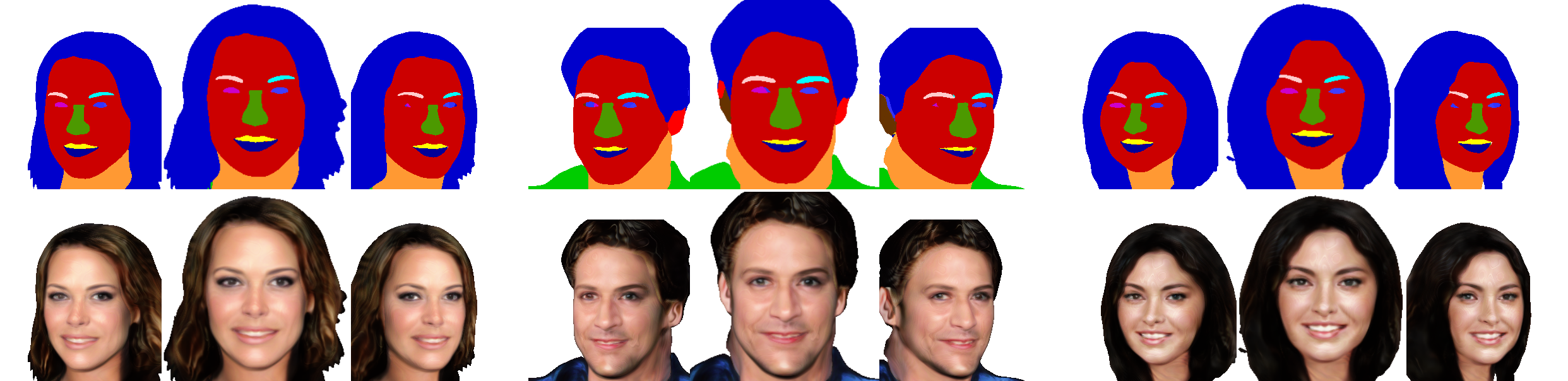}
\end{center}
    \vspace{-0.4 cm}
    \caption{
    \textcolor{black}{Renderings of Semantic-radiance and RGB-radiance on image space generated by our approach by random latent code $\mathbf{z} \in \mathrm{R}^d$ and $\mathbf{z}_i \in  \mathrm{R}^d$. 
    }}
    \label{fig:samples}
\end{figure}

\section{Related Works}


\noindent \textbf{Neural Implicit Representations.} Neural Implicit functions have been widely applied to various vision problems such as novel view synthesis \cite{tucker2020single,sitzmann2019deepvoxels}, and image manipulations \cite{abdal2021styleflow,zhou2021cips}. 
Furthermore, they have gained significant traction in diverse 3D-related tasks, including 3D scene reconstruction \cite{mildenhall2021nerf}, and occupancy field or signed distance function estimation \cite{or2022stylesdf,michalkiewicz2019implicit}.
Recent works have explored alternative representations of implicit functions, such as octrees and tri-planes \cite{chan2022efficient,yu2021plenoctrees}, for faster inference and improved expressive power. 
Additionally, GRAM \cite{deng2022gram} introduces an effective point sampling method and radiance field learning on 2D manifolds, enabling fine detailed learning with reduced computational complexity.
This paper emphasizes learning semantic information on 2D manifolds through a geometry module rather than directly learning about the image space. 
However, learning the semantic field alone encounters challenges due to the absence of depth information in the semantic mask data. 
To address this, the learning of the semantic field is further supervised by an additional appearance module. 
Unlike \cite{sun2022fenerf} and \cite{jiang2022nerffaceediting}, our appearance module uses the semantic information from the geometry module to control the appearance of each semantic point on these manifolds in the volume.

\noindent \textbf{3D-aware Neural Face Image Synthesis.} 
The combination of Generative Adversarial Networks (GANs), as originally proposed  \cite{goodfellow2014generative}, and neural implicit radiance fields (NeRF) \cite{mildenhall2021nerf}, has been widely explored for the generation of 3D-consistent faces. 
Despite utilizing 3D aware features and 2D CNN-based renderers to generate synthetic images, several methods such as \cite{gu2021stylenerf,liao2020towards,niemeyer2021giraffe,chan2021pi} face challenges in accurately representing geometry, resulting in lower quality representations.
Recent studies \cite{an2023panohead,deng2022gram,chan2022efficient,or2022stylesdf} investigate novel representations and models, enabling the synthesis of geometrically consistent fine details. 

\noindent \textbf{Editing Faces with 2D GANs \textcolor{black}{ and DDPMs.}}
Generative adversarial networks (GANs)\textcolor{black}{, and more recently, Denoising Diffusion Probabilistic Models (DDPMs) have widely been applied} for generating realistic images, with face editing as a prominent area of research in this context. 
Various methods have been proposed for face editing using semantic maps. \textcolor{black}{These approaches leverage conditional 2D GANs \cite{park2019semantic,zhu2020sean,chen2022sofgan,leimkuhler2021freestylegan}, and DDPMs \cite{huang2023collaborative,kim2022diffusionclip,sun2022anyface} which are conditioned on semantic masks.}
Unlike explicit field/occupancy-based neural scene representations, SofGAN \cite{chen2022sofgan} utilizes implicit fields for semantic occupancy generation. However, SofGAN relies on semantically labelled 3D scans for training and involves a separate process to synthesize images through 2D projections of semantic labels. 
\textcolor{black}{Diffusion-Rig \cite{ding2023diffusionrig} first learn facial priors by fitting a morphable face model for conditioning diffusion model to generate view-consistent 2D facial images.
In contrast, our proposed framework directly modifies appearance in the generated implicit neural representation without the need for 2D projections or facial priors, and is trained on a collection of 2D images instead of 3D scans.}

\noindent \textbf{Editing Faces in Neural Radiance Fields.} 
To enable control over the geometry and appearance of synthesized images by 3D-aware GANs, subsequent works employ the embedding of explicit control into the generation process. Certain methods, including \cite{athar2023flame,zhuang2022mofanerf,zheng2022avatar}, condition radiance fields with information from 3D Morphable models (3DMM) \cite{li2017learning,paysan20093d} to exert control over the generated fields. 
This conditioning is achieved through embeddings of latent spaces representing pose, shape, and expression parameters of the 3DMM models.
FENeRF \cite{sun2022fenerf} utilizes semantic masks for geometry control and a separate latent vector for appearance control in synthesized images. 
CG-NeRF \cite{jo2021cg} introduces soft conditions, including sketches as external inputs, to influence radiance field generation. 
Following \cite{chan2022efficient}, IDE-3D \cite{sun2022ide} enables editing on tri-plane representations while NeRFFaceEditing \cite{jiang2022nerffaceediting} disentangles geometry and appearance using pre-trained tri-plane representations, eliminating the need for retraining. 

While the aforementioned methods provide control over the geometry and appearance of synthesized images, they fall short in offering semantic-specific control. 
Our proposed method provides seamless semantic-specific control over geometry and appearance attributes in facial images, as shown in Figure \ref{fig:gedit}. This allows for effortlessly transferring specific facial attributes to other images, including overall geometry and appearance.



\section{Methodology}
Our primary objective is to learn semantic information, occupancy, and colour attributes for points sampled on 2D Manifolds. The Manifold Predictor model predicts these manifolds \cite{deng2022gram}. 
To separate the representation of geometry and texture during the image generation process, we introduce two distinct modules: the Geometry module and the Appearance module, as illustrated in Figure \ref{fig:arch}. 
The Geometry module employs a shape latent code for each point on the manifolds, influencing occupancy and semantic information. Simultaneously, it predicts high-dimensional Appearance Descriptors $\mathcal{F}$ for each point. 
These descriptors are segregated based on semantic information using the Semantic Volume Masking Layer. 
The Appearance module then conditions points belonging to a specific semantic with different appearance latent codes and predicts colour information. 
The entire method is trained in two stages using adversarial learning. Initially, we train the entire network end-to-end, simultaneously learning weights for the Geometry and Appearance modules. Later, we freeze the weights of the Geometry module and fine-tune the Appearance module to achieve high-quality volumetric renderings of generated images.
\subsection{Network Architecture}
Our generator consists of the Geometry Module and the Appearance Module, as illustrated in Figure \ref{fig:arch}. Both modules are implemented as Multi-Layer Perceptrons (MLPs) to produce occupancy $\sigma$, semantic radiance $s$, and colour $c = (r,g,b)$ values for a given point $x \in \mathrm{R}^3$ with view direction $d \in \mathrm{R}^3$.
\begin{equation}
    \Psi_g:(\mathbf{z},\mathbf{x},\mathbf{d}) \mapsto (\mathbf{\sigma},\mathbf{s}), \:
    \Psi_a:(\{\mathbf{z_i}\}_{i=0}^{l},\mathbf{x},\mathbf{d},\mathbf{s}) \mapsto (\mathbf{c})
    \label{eq:geneq}
\end{equation}

where, $\mathbf{z} \in \mathrm{R}^d \sim p_z$ represents the geometric latent code and $\mathbf{z_i} \in \mathrm{R}^d \sim p_z$ represents the appearance latent code for each semantic labels $l$. 

\noindent \textbf{Manifold Predictor.} 
We sample $N$ points along each ray cast in the direction $\mathbf{d}$, given a camera pose $\theta \in \mathrm{R}^3 \sim p_\theta$. 
The Manifold Predictor $\mathcal{M}$ is implemented as a scalar field function using light-weight MLPs, predicting a scalar value $s$ for the sampled points. 
From the predicted scalar field, we extract $K$ iso-surfaces with different levels ${l_i}$ and perform final point sampling by finding the nearest point of intersection on these iso-surfaces, following \cite{deng2022gram}. 

\begin{figure}[t]
\begin{center}
    \includegraphics[width=\linewidth]{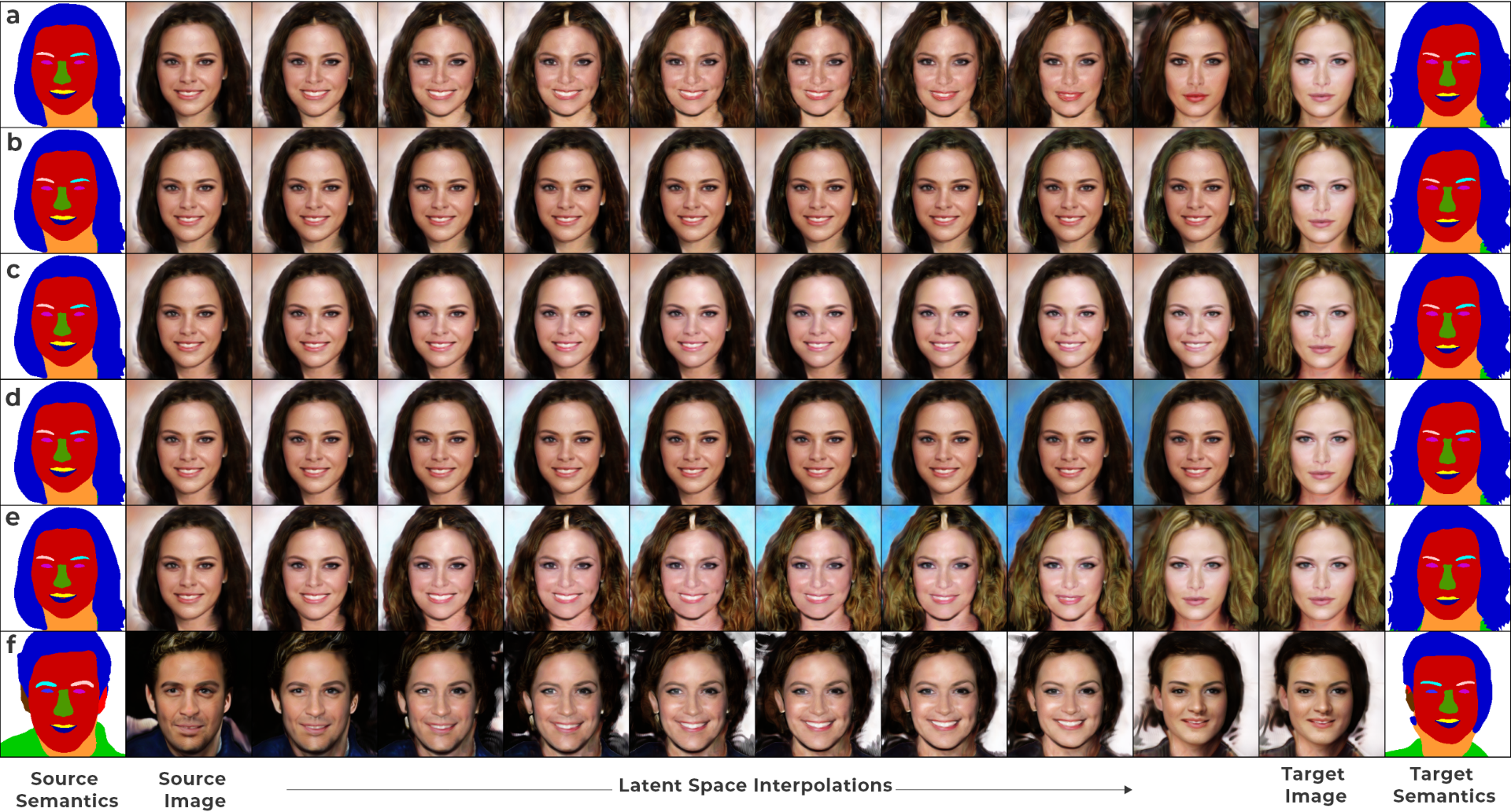}
\end{center}
    \vspace{-0.4cm}
    \caption{Latent space disentanglement and interpolations. 
     In (a), linear interpolation is performed on the geometric latent space $\mathrm{z}$ while keeping all appearance latent $\mathrm{z}_i$ fixed. (b-d) demonstrate changes in appearance latent variables corresponding to hair, face, and background semantics. All latent codes are linearly interpolated for (e) and (f). Additionally, (f) showcases the change in the appearance of garment semantics from the source to the target.
    }
    \vspace{-0.2cm}
    \label{fig:interp}
\end{figure}

\noindent \textbf{Geometry Module.} 
Given a latent code $\mathrm{z}$, the Geometry module $\Psi_g$ generates the radiance for each point on the predicted 2D manifolds in volume. The architecture of $\Psi_g$ consists of FiLM SIREN backbone, inspired from \cite{chan2021pi} with some modifications, as presented in Figure \ref{fig:archga}. Instead of directly synthesizing images as in \cite{deng2022gram}, our geometric module first learns to predict occupancy $\sigma$ and semantic radiance field, $\mathbf{s}$. 
Our primary goal is to disentangle geometry from appearance by learning geometry independently of image synthesis. 
However, attempting to learn semantic radiance solely using semantic masks in an adversarial setting is bound to fail, as the perceptual appearance of a semantic mask lacks depth information present in images due to the absence of shading effects caused by varying shades of light.

To address this limitation, we additionally predict high-dimensional feature descriptors $\mathcal{F}$ for each point, which is then conditioned by the Appearance module for image synthesis. By backpropagating gradients through $\mathcal{F}$ from the appearance module into the geometry module, we ensure the correct learning of the occupancy field $\sigma$. This approach helps in better disentangling geometry and appearance, leading to improved results in our framework.

\noindent \textbf{Semantic Volume Masking Layer.}
Given Semantic radiance field $\mathbf{s}$ and occupancy field $\mathbf{\sigma}$, we perform volumetric rendering \cite{oechsle2021unisurf,zhou2018stereo} on all 2D manifolds. When a ray $\mathrm{r}$ is cast and intersects $K$ surface manifolds at points ${\mathrm{x}_j}$ sorted by proximity, we can describe the semantics of a point on the $k$-th manifold out of the total $K$ manifold:
\begin{equation}
 \mathrm{S}_k(r) = \arg \max \sum_{j=k}^{K} T(\mathrm{x}_j) \alpha(\mathrm{x}_j)\mathrm{s}(\mathrm{x}_j, \mathrm{d}), \;\; \mathbf{for} \;\; k \leq K 
 \label{eq:vr}
\end{equation}
Here, $T(x_j) = \prod_{i<j; i \geq k} (1 - \mathbf{\sigma}(\mathrm{x}_j))$, and Equation \ref{eq:vr} allows us to classify each sampled point to the semantic class it belongs to.
After obtaining the semantic labels, we segregate the points and their corresponding appearance descriptors $\mathcal{F}$ based on their specific semantic categories. Using the Appearance module, this segregation allows us to condition them on respective semantic-specific latent codes.

Let there be $n$ semantic classes, and the Semantic Volume Masking layer segregates all sampled points in the volume into $n$ collections, denoted by $\mathrm{x_i}$ along with their corresponding casted direction $\mathrm{d_i}$ and appearance descriptor $\mathcal{F}_i$. These collections are created such that every point $ x \in \mathrm{x}_i$ belongs to a specific semantic category $i$, defined among the $n$ available semantic classes.

\noindent \textbf{Appearance Module.}
The architecture of our appearance module closely resembles that of the Geometry Module, as depicted in Figure \ref{fig:archga}. 
However, it is distinguished by the presence of $n$ mapping networks, each responsible for conditioning the $i$-th collection obtained from the Volume Masking layer. 
These mapping networks are utilized to condition the Appearance Descriptors $\mathcal{F}_i$, which then pass through a SIREN-based neural network for RGB radiance prediction. 
As a result, our appearance module $\Psi_a$ generates a radiance field $\mathrm{c}$, where the $i$-th latent code controls the appearance of the $i$-th semantic among all $n$ semantic classes.
Importantly, it should be noted that while different mapping networks are used for conditioning the collections, the weights of the SIREN backbone in the appearance module remain consistent across all collections.
In the Results Section, we present exhaustive ablation studies conducted on the Appearance Module to identify the optimal network architecture. The Appearance Module takes segregated Appearance Descriptors from the geometry module and a latent code for each semantic as input, producing the final RGB radiance. Subsequently, we perform volumetric radiance integration with the occupancy field $\sigma$, obtained from the Geometry Module, following a similar approach as in \cite{oechsle2021unisurf,deng2022gram}, to obtain the final image.

\section{Experiments and Results}

\subsection{Training}
 We train SemFaceEdit on the CelebAMask-HQ \cite{CelebAMask-HQ} dataset. 
The dataset contains 30K high-resolution face images and annotated masks in $N=19$ classes. 
In our experiments, we club $N$ classes in 4 semantics, denoted as $n$: hairs, face (containing eyes, nose, mouth, neck, and ears), garment, and background. 
We train our network in two stages in an adversarial learning setting, using non-saturating GAN loss with R1 regularization \cite{mescheder2018training}. 
During training, we randomly sample geometric latent code $\mathrm{z}$, $n$ appearance latent codes $\mathrm{z}_i$ and pose $\theta$ from Gaussian distributions $p_z^g, p_{z_i}^a$, and $p_\theta$, respectively. 
The generator, consisting of $\Psi_g$ and $\Psi_a$, produces semantic $\mathrm{s}$, occupancy $\sigma$, and RGB $\mathrm{c}$, radiance fields. 
In the first stage, we jointly train $\Psi_g$ and $\Psi_a$ with two discriminators, Semantic map discriminator $D_s$ and Facial image discriminator $D_c$. \textcolor{black}{We provide further details on data preprocessing, discriminator architecture and loss functions in \textit{Suppl. material}.}
\textcolor{black}{In Figure \ref{fig:samples}, We present multi-view images generated from randomly sampled latent codes after training the model.}

\begin{figure}[t]
\begin{center}
    \includegraphics[width=\linewidth]{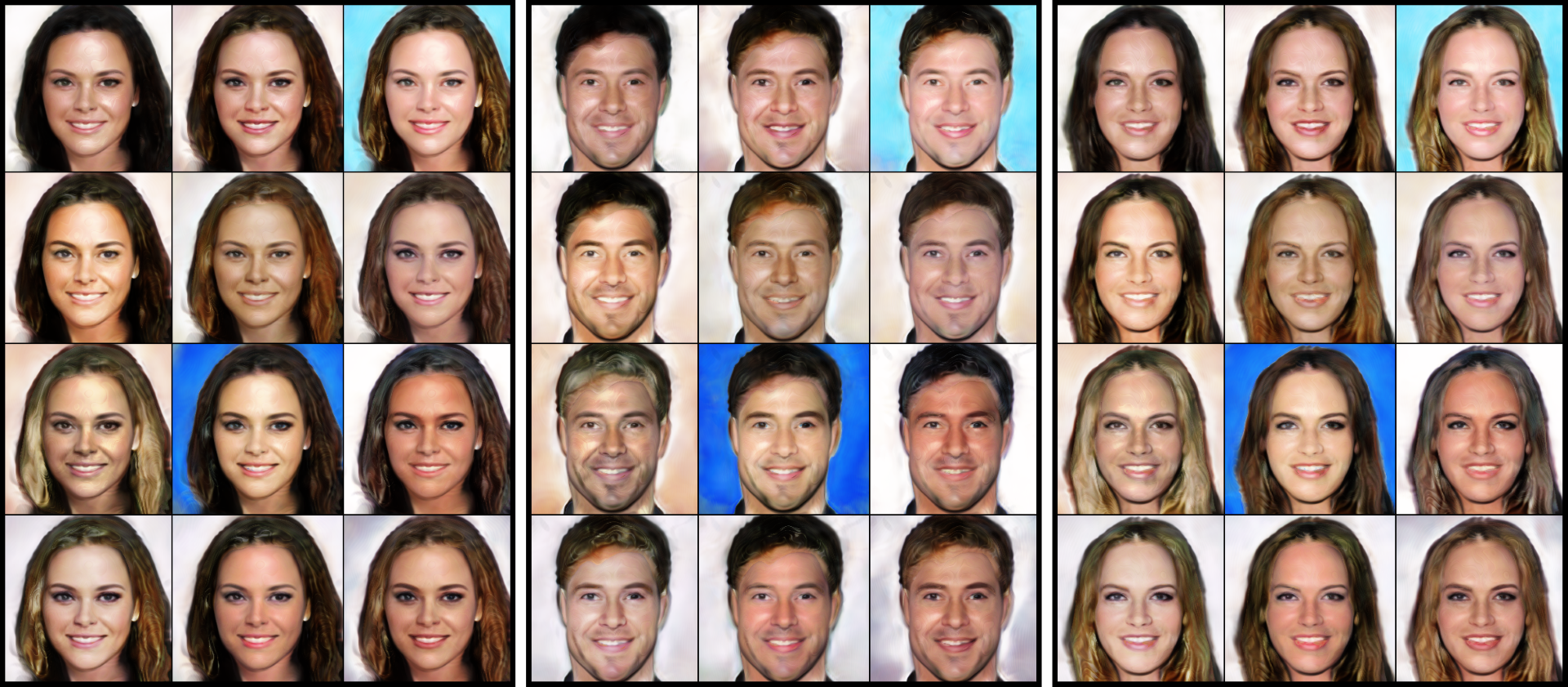}
\end{center}
    \caption{
    \textcolor{black}{Appearance Latent Codes ($z_i$) influence on Geometric Shapes. The image (i,j) in grid (a),(b), and (c) are generated with same $z_i$'s, highlighting resemblance in appearance across hair, background, and facial regions with (i,j)$^{th}$ image in each grid.}
    }
    \label{fig:app}
    \vspace{-0.3cm}
\end{figure}

\noindent \textbf{Implementation Details.} 
Our experiments utilize the Adam Optimizer \cite{kingma2014adam} with learning rates of $2 \times 10^{-5}$ for the generator and $2 \times 10^{-4}$ for the discriminator. 
Utilizing 2 NVIDIA GeForce RTX 4090 GPUs, each with 24 GB of memory, the model is trained for 120K iterations (3 epochs) during the first stage, followed by 30K iterations (1 epoch) in the second stage. \textcolor{black}{
To generate neural rendering of radiance field on to image space $\mathcal{I} \in \mathrm{R}^{W \times H \times 3}$, we cast $W \times H$ rays from the camera and sample $K$ number of points along each ray, where $K=24$ is number of iso-surfaces on manifolds. Specefically, we sample radiance values of 3.14 million ($256 \times 256 \times 24 \times 2$) points in total from semantic-radiance field $s$ and rgb-radiance field $c$. To overcome computational challenges, we utilize PyTorch’s Automatic Mixed Precision (AMP) to reduce memory cost. During inference, our approach generates $256^2$ and $128^2$ image in $0.86$ and $0.15$ seconds with peak GPU memory usage of approx. $15$GB and $6$GB, respectively.
}

\begin{figure}[t]
\begin{center}
    \includegraphics[width=\linewidth]{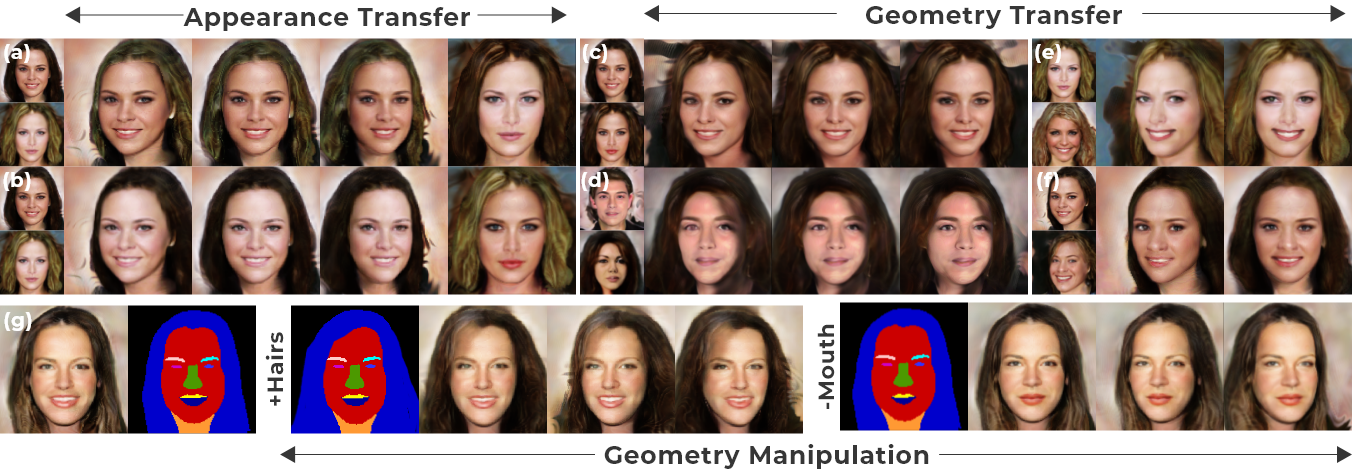}
\end{center}
    \vspace{-0.4cm}
    \caption{
     (a-f) displays two smaller images, with the top image representing the source attribute and the bottom image representing the target attribute of appearance/geometry. 
    \textbf{Appearance Transfer}: In (a), we present three different views of the source image showcasing the transferred hairstyle appearance from the target to the source image. Additionally, we present a single view of a converse transfer of appearance, i.e. from the source to the target image. Similarly, in (b), we showcase the transferred facial appearance.
    \textbf{Geometry Transfer}: (c) and (d) demonstrate the transfer of target hair geometry to the source image. In contrast, (e) and (f) showcase the smile and nose geometry transfer from the target to the source image. Note that in (c) and (d), alterations in attributes at regions other than hair semantics are minimal. In contrast, the transferred hair adopts the appearance style of the target image through manipulation of the latent code governing hair semantics.
    The focus is solely on transferring geometry for finer semantics like the nose or smile in (e, f). Therefore, we constrain appearance transfer by maintaining the unchanged state of appearance latents.
    \textbf{Geometry Manipulation}: (g) Shows the effect of manipulation in the semantic mask by expanding the hair region semantics and the effect of shrinking at the mouth region.
    }
    \label{fig:gedit}
    \vspace{-0.2cm}
\end{figure}
 
\subsection{Editing Faces in Radiance Fields.}
\begin{figure}[t]
\begin{center}
    \includegraphics[width=0.95\linewidth]{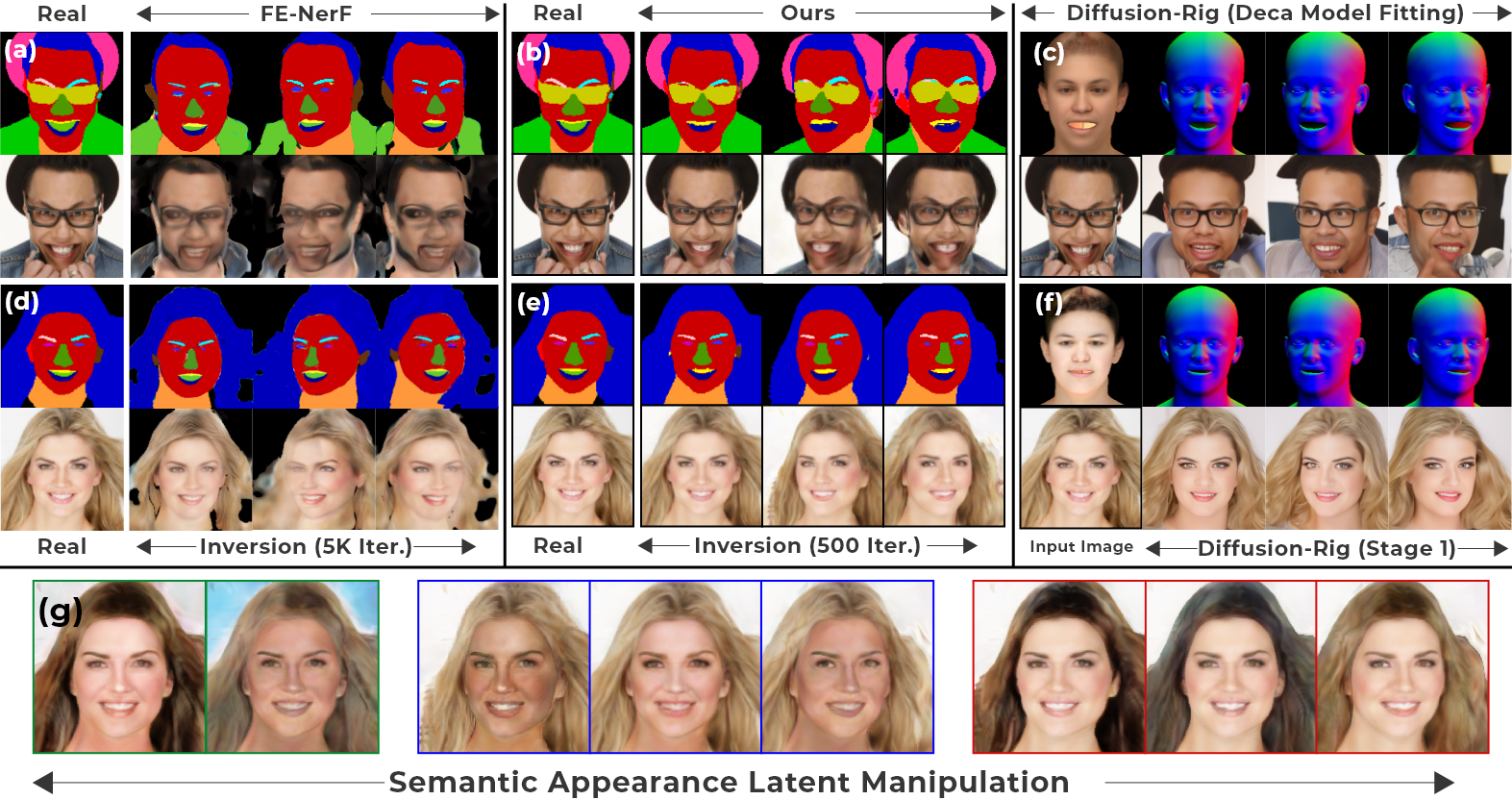}
\end{center}
    \vspace{-0.3cm}
    \caption{
    Comparison of pivotal inversion results between our proposed approach (b,e) and FE-NerF \cite{sun2022fenerf} (a,d). Our method achieves rapid convergence within 500 iterations, contrasting with FE-NerF's 5K iterations. \textcolor{black}{Additionally, we compare with Diffusion-Rig \cite{ding2023diffusionrig} (c,f), which first estimate facial prior by fitting a Deca model \cite{feng2021learning} on given image. Subsequently it employs a diffusion-based approach to generate view consistent facial images.} (g) demonstrates semantic appearance variation achieved by modifying latent codes specific to distinct semantic regions $\mathbf{z_i}$ while maintaining geometry latent $\mathbf{z}$ of (d). Green outline signifies full appearance alteration; blue outlines indicate facial appearance changes and red outlines represent hair appearance modifications. Real images for inversion are taken from CelebAMask-HQ Dataset \cite{CelebAMask-HQ}.
    }
    \label{fig:sota}
    \vspace{-0.2cm}
\end{figure}

To edit an input image obtained either from $\mathrm{I} \sim p_{rim}$ (real images) or synthetically generated using Generators $\Psi_g$ and $\Psi_a$, we perform image inversion into the $\mathcal{W}$ space, represented as $\textit{w}$, utilizing pivotal tuning inversion \cite{roich2022pivotal}.
To enable image editing based on semantic mask manipulations, we require the original mask $S$ and the user-edited mask $S'$. For real face images, we utilize the method proposed by \cite{yu2018bisenet} to obtain semantic mask $S$. On the other hand, for synthetic images generated through our pipeline, the Geometry module $\Psi_g$ naturally provides us with the required semantic mask. 
Our experiments utilize trained mapping networks to obtain the average frequency and phase shifts over 10k sampled latent codes for geometry $z$ and appearance $z_i$. For brevity, we denote these averaged frequencies and phase shifts as a latent code $\textit{w} \in \mathcal{W}$ and $\textit{w}_i \in \mathcal{W}_i$ for geometry and appearance, respectively.
To achieve editing, we optimize an editing offset vector $\delta\textit{w}^+ \in \mathcal{W}$ such that the generated semantic mask $S'$ from $\Psi_g$ approximates the original mask $S$. During optimization, we keep $\textit{w}_i$ for $\Psi_a$ fixed while optimizing for the geometric latent $\delta\textit{w}^+$. We directly employ the corresponding $z_i$'s for the appearance of synthetically generated images. For real images, we further optimize for $\delta \textit{w}_i^+$ to make the appearance resemble the given image $\mathrm{I}$.
Figure \ref{fig:gedit} showcases editing in geometry based on semantic masks, and Figure \ref{fig:sota}(b,d) demonstrates inversion and latent code manipulations on real images from CelebAMask-HQ Dataset.



\noindent \textbf{Geometry and Appearance Transfer}.
We can directly synthesize appearance manipulations targeting a specific semantic $k \in n$ using $\Psi_a$. Unlike \cite{jiang2022nerffaceediting}, $\Psi_a$ controls the change in appearance of regions corresponding only to the $k$-th semantic. We illustrate these results qualitatively in Figures \ref{fig:app} and \ref{fig:interp}. However, as discussed previously, image inversion for Geometry transfer into the $\mathcal{W}$ space is required.
Given images $\mathrm{I}_1$ and $\mathrm{I}_2$ with semantics $S_1$ and $S_2$, respectively, let $(\textit{w}^1, \textit{w}^2)$ represent geometric latents and $(\textit{w}^1_i, \textit{w}^2_i)$ be appearance latent codes. We want to transfer the semantic region $k \in n$, represented by $\mathcal{M}_k$, from image $\mathrm{I}_2$ to $\mathrm{I}_1$. For appearance transfer, we directly interchange the $k$-th appearance semantic latent code in $\textit{w}_i^1$ with $\textit{w}_k^2$, and then optimize for $\delta\textit{w}^+$, such that $\Psi_g(\textit{w}^1 + \delta\textit{w}^+, \theta)$ approximates the target semantic $S_1 \odot (1-\mathcal{M}_k) + S_2 \odot \mathcal{M}_k$, and $\Psi_a(\textit{w}^1_i, \theta)$ approximates the target image $\mathrm{I}_1 \odot (1-\mathcal{M}_k) + \mathrm{I}_2 \odot \mathcal{M}_k$. 
Though $\textit{w}^1_i$ for appearance is fixed, inversion requires a backward pass through the $\Psi_g$ and $\Psi_a$ to update $\delta\textit{w}^+$. This is because $\Psi_g$ depends on $\Psi_a$ for the correct estimation of appearance feature descriptors $\mathcal{F}$ and occupancy $\sigma$.
In Figure \ref{fig:gedit} (c-g), we show the geometry transfer from target to source of various semantics. More details can be found in the \textit{Suppl. Material}.


\begin{figure}[t]
\begin{center}
    \includegraphics[width=\linewidth]{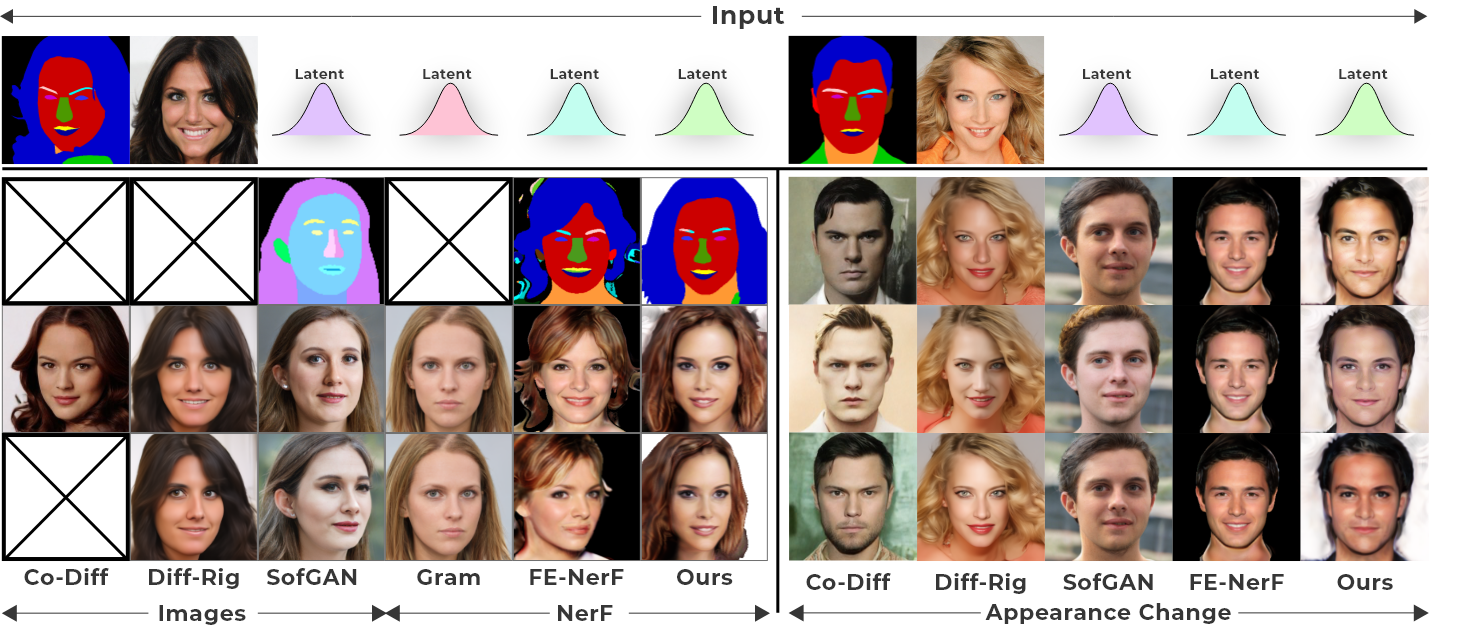}
\end{center}
    \vspace{-0.3cm}
    \caption{
    \textcolor{black}{Qualitative comparison of SemFaceEdit with Co-Diff \cite{huang2023collaborative}, Diff-Rig \cite{ding2023diffusionrig}, SofGAN \cite{chen2022sofgan}, GRAM \cite{deng2022gram}, and FE-NerF \cite{sun2022fenerf}.
    $\boxtimes$ represents that the respective approach don't support or yield the required result. 
    Input mask in \cite{huang2023collaborative} and image in \cite{ding2023diffusionrig} are taken from CelebAMask-HQ Dataset\cite{CelebAMask-HQ}. 
    }}
    \label{fig:sota_geoapp}
    \vspace{-0.4cm}
\end{figure}

\subsection{Results}
\textcolor{black}{We conduct a comparative analysis of SemFaceEdit against five state-of-the-art techniques: Co-Diff \cite{huang2023collaborative}, Diff-Rig \cite{ding2023diffusionrig}, GRAM \cite{deng2022gram}, FE-NerF \cite{sun2022fenerf}, and SofGAN \cite{chen2022sofgan}. GRAM and FE-NerF are oriented towards generating 3D-aware images, while SofGAN employs a 2D projection of the 3D semantic occupancy field for dynamic styling through a 2D GAN methodology.
In contrast to SofGAN, Diff-Rig \cite{ding2023diffusionrig} employs a diffusion-based approach for view-consistent 2D image generation. This method lacks semantic-guided editing capability, instead utilises expression basis and priors from a morphable face model for appearance changes.
We also include Co-Diff \cite{huang2023collaborative}, a 2D facial image generation approach, which generates realistic facial images given a semantic map but lacks multiple view generation.}

\noindent \textbf{Qualitative comparison}. 
Our framework estimates semantic radiance in addition to the occupancy field and RGB radiance. 
In Figure \ref{fig:sota_geoapp} we showcase the generated semantic maps and multi-view images produced by our method compared to the aforementioned approaches. Additionally, on the right, we display the appearance manipulation achieved by our method in comparison with the other techniques.
\textcolor{black}{Co-Diff takes a semantic map as input and generates a facial image that fits the provided map, but it cannot generate multi-view images.
Diff-Rig first estimates facial priors by fitting Deca model \cite{feng2021learning} on the given image, then employs a diffusion-based approach to generate view-consistent facial images in 2D space. Changing these buffers along with diffusion noise allows for appearance manipulation and editing, but this methodology lacks semantic-guided editing by design.}
Sof-GAN produces high-resolution, multi-view facial appearances using a 2D GAN approach, but its training relies on semantically labeled 3D scans. However, the use of 2D GANs prevents it from achieving 3D interpretability.
The method presented in GRAM introduces an effective point sampling technique on learnable 2D manifolds. Nonetheless, it falls short in estimating semantic radiance and achieving disentanglement between geometry and appearance. 
FE-NerF predicts the semantic field but falls short in terms of disentanglement and control over semantic appearance.

Our 3D inversion outcomes, showcased in Figure \ref{fig:sota}(b,e), are compared against FE-NerF depicted in Figure \ref{fig:sota}(a,d), respectively. Notably, the semantic inversion by FE-NerF struggles with intricate semantics and produces noisy results, as demonstrated in Figure \ref{fig:sota}(a). In contrast, our method achieves successful 3D inversion for such cases. Our approach further enables semantic appearance control on the generated RGB-radiance fields.
\textcolor{black}{After obtaining optimized geometry latent $\mathbf{z}$ and appearance latents $\mathbf{z}_i$ through inversion process, we can directly manipulate appearances by changing latent codes $\mathbf{z}_i$ corresponding to a semantic region $i$, as demonstrated in Figure \ref{fig:sota}(g).
In contrast to our approach which relies on pivotal inversion, Diff-Rig generate the images, shown in Figure \ref{fig:sota}(c,f) by first estimating facial priors by fitting Deca face model \cite{feng2021learning} on the input image.  Note that for Diff-Rig, we only present the output generated after Stage-1 in Figure \ref{fig:sota}. More accurate results can be obtained by fine-tuning the model on a personalized album, however, this requires multi-view and diverse images of the same face.}

\begin{table}[t]
\caption{Comparing FID and KID between 10K generated and 10K real CelebA-MaskHQ \cite{CelebAMask-HQ} Dataset images with GRAM \cite{deng2022gram}, FE-NerF \cite{sun2022fenerf}, SofGAN \cite{chen2022sofgan}, \textcolor{black}{Diffusion-Rig \cite{ding2023diffusionrig}, and Co-Diff \cite{huang2023collaborative}}.  Lower values ($\downarrow$) indicate better performance. \textcolor{black}{Superfixes $^*$ and $^\dag$ indicates that methods do not generate neural rgb-radiance fields but employ 2-dimensional deep neural networks (GANs/DDPMs) to synthesize multi-view consistent images ($^*$) or only a single view image ($^\dag$).}
Abbreviations used: GAC - Global Appearance Control, LAC - Local Appearance Control, and SGE - Semantic Guided Editing. \textcolor{black}{Note that Diffusion-Rig do not support SGE but can perform editing utilizing expression basis and physical buffers of fitted morphable DECA model.}}
\vspace{-0.1cm}
\begin{center}
\scriptsize
\resizebox{\linewidth}{!}{%
\begin{tabular}{l|c|c|c|c|c|c|c|r}
\hline
\multicolumn{1}{c|}{Methods} &
  \multicolumn{1}{c|}{\begin{tabular}[c]{@{}c@{}}3D RGB \\ Radiance\end{tabular}} &
  \multicolumn{1}{c|}{\begin{tabular}[c]{@{}c@{}}3D Semantic \\ Radiance\end{tabular}} &
  \multicolumn{1}{c|}{\begin{tabular}[c]{@{}c@{}}GAC\end{tabular}} &
  \multicolumn{1}{c|}{\begin{tabular}[c]{@{}c@{}}LAC\end{tabular}} &
  \multicolumn{1}{c|}{\begin{tabular}[c]{@{}c@{}}SGE\end{tabular}} &
  \multicolumn{1}{c|}{\begin{tabular}[c]{@{}c@{}}FID $\downarrow$ \\ $(256^2)$\end{tabular}} &
  \multicolumn{1}{c|}{\begin{tabular}[c]{@{}c@{}}KID $(\times 10^3)$ $\downarrow$ \\ $(256^2)$\end{tabular}} 
  &
  \multicolumn{1}{c}{\begin{tabular}[c]{@{}c@{}}\textcolor{black}{\#Pars}\end{tabular}} \\ \hline
GRAM    & \checkmark & \xmark  & \xmark & \xmark & \xmark & 17.36 & 30.82 & \textcolor{black}{1.95 M}\\ \hline
FE-NerF & \checkmark & \checkmark & \checkmark & \xmark & \checkmark & 28.97 & 39.33 & \textcolor{black}{31.09 M}\\ \hline
SemFaceEdit (Ours)    & \checkmark & \checkmark & \checkmark & \checkmark & \checkmark & 19.81 & 32.81 & \textcolor{black}{8.78 M}\\ \hline
\begin{tabular}[c]{@{}l@{}}SemFaceEdit (Ours)\\ (Single Stage)\end{tabular}   & \checkmark & \checkmark & \checkmark & \checkmark & \checkmark & 21.27 & 34.73 & \textcolor{black}{8.78 M}\\ \hline
\hline
SofGAN*  & \xmark     & \checkmark & \checkmark & \checkmark & \checkmark & 25.72 & 30.86 & \textcolor{black}{30.68 M}\\ \hline
\textcolor{black}{Diffusion-Rig*} & \textcolor{black}{\xmark}     & \textcolor{black}{\xmark} & \textcolor{black}{\checkmark} & \textcolor{black}{\xmark} & \textcolor{black}{\xmark} & \textcolor{black}{16.84} & \textcolor{black}{30.64} & \textcolor{black}{144.6 M}\\ \hline
\textcolor{black}{Co-Diff$^\dag$}  & \textcolor{black}{\xmark}     & \textcolor{black}{\xmark} & \textcolor{black}{\checkmark} & \textcolor{black}{\xmark} & \textcolor{black}{\checkmark} & \textcolor{black}{14.63} & \textcolor{black}{27.43} & \textcolor{black}{459.6 M} \\ \hline
\end{tabular}%
}
\end{center}
\label{tab:quant}
\vspace{-0.3cm}
\end{table}

\begin{figure}[t]
\begin{center}
    \includegraphics[width=\linewidth]{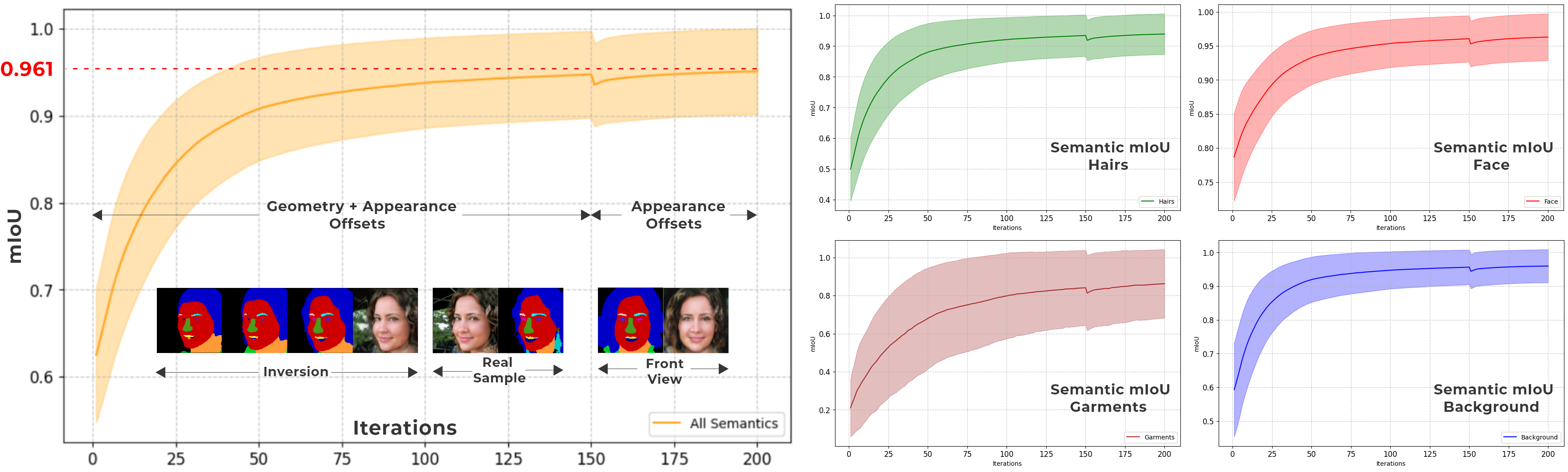}
\end{center}
    \vspace{-0.2cm}
    \caption{
    During the optimization process for 3D inversion on 1000 CelebAMask-HQ Dataset images, we track the Mean Intersection over Union (mIoU). Embedded images qualitatively illustrate the inversion progression of an image in a challenging pose. Furthermore, we track the Mean Intersection over Union (mIoU) of all $n=4$ semantics, namely: Hairs, Face, Garments, and Background.
    }
    \label{fig:mIoU}
    \vspace{-0.2cm}
\end{figure}

\noindent \textbf{Quantitative comparison}.
Table \ref{tab:quant} presents the assessment of image quality using the Frechet Inception Distance (FID) \cite{heusel2017gans} and Kernel Inception Distance (KID) \cite{binkowski2018demystifying}. This evaluation is performed between 10K generated images from randomly sampled latent codes and an equal number of randomly selected real images. 
We significantly improve the neural rgb-radiance generation compared to that of FE-NerF, as evident in Table \ref{tab:quant} while jointly predicting the semantic radiance field. 
\textcolor{black}{For Diffusion-Rig, we use 10K real images from the CelebAMask-HQ dataset. We estimate buffers (pose, normal, and albedo) by fitting the Deca Model \cite{feng2021learning} and generate outputs from stage-1, as proposed by \cite{ding2023diffusionrig}. In contrast, for Co-Diff, we input 10K semantic masks from the CelebAMask-HQ dataset for facial image generation to calculate metrics.}
To assess the 3D inversion capability of SemFaceEdit, we perform inversions on 1K images from the CelebAMask-HQ Dataset, calculating the mean Intersection over Union (mIoU) for all $n$ semantics. 
Across all inverted images, our approach converges to 0.85 mIoU within 50 iterations. The per-iteration mIoU and semantic class-wise mIoU progressions are depicted in Figure \ref{fig:mIoU}.

\noindent \textbf{Ablation Studies.}
We conducted several ablation studies to determine the optimal choice for SemFaceEdit architecture. In Table \ref{tab:quant}, we show improved generated image metrics after second-stage training. This process involves refining Appearance Module weights while keeping Geometry module weights frozen. 
The \textit{Suppl. Material} contains extra ablation studies examining (1) the impact of shared Appearance Modules and (2) Depth Ablation.

\noindent \textcolor{black}{\textbf{Limitations.}
Our appearance module uses the generated semantic radiance from the geometry module through a semantic volume masking layer. The generation of RGB-radiance is constrained to adhere to the densities of the semantic radiance, which limits the creation of detailed and fine-grained geometry. Additionally, any discrepancies in the semantic geometry propagates to final facial appearance.  We include illustrations of such cases in \textit{supp. material}. The high computational requirements for training generative NeRFs limit the proposed method's ability to generate high-quality images, resulting in image quality that lags behind traditional methods using 2D GANs and DDPMs. In the future, more efficient point sampling and implicit representations could be explored to reduce computational overhead and achieve higher quality radiances.
}

\section{Conclusion}
This work introduces a novel approach for semantic face editing on generative radiance manifolds. We enable effective editing capabilities by simultaneously estimating each point's semantic information and RGB radiance.   
The key to our approach lies in utilizing a differentiable Semantic Volume Masking layer, which effectively partitions points according to their predicted semantics.  This partitioning allows us to condition the RGB radiance with latent codes, resulting in a finely tuned level of control over the editing process. With SemFaceEdit, we successfully attain comprehensive control over geometry and appearance, including transferring distinct facial attributes between images. Through experiments, we demonstrate the capability of our method to effectively control the visual attributes of distinct semantics while preserving other regions by manipulating latent codes. Moreover, SemFaceEdit achieves enhanced GAN inversion and faster convergence compared to existing state-of-the-art methods.

%
%
%
%

{
    \small
    \bibliographystyle{splncs04}
    \bibliography{main}
}

\clearpage
\section*{\centering Appendix}
\setcounter{section}{0}
\section{Loss Functions}
\textcolor{black}{
As discussed in the main paper, we train our network in two stages in an adversarial learning setting, using non-saturating GAN loss with R1 regularization \cite{mescheder2018training}. 
In the first stage, we jointly train $\Psi_g$ and $\Psi_a$ with two discriminators, Semantic map Discriminator $D_s$ and Facial image discriminator $D_c$, using the loss function given in Equation \ref{eq:s1loss}. 
Our discriminators have similar architectures to those utilized by \cite{chan2021pi}, with minor changes. 
To ensure that the final images generated by the Appearance Module $\Psi_a$ are in the correct pose, we introduce an additional branch consisting of linear layer within discriminator $D_c$ to estimate the pose $\theta$. We also add an additional branch consisting of linear layer within $D_s$ to estimate latent code that generated the semantic radiance. The adversarial losses for discriminating semantics and images is defined in Equation \ref{eq:loss}
}

\begin{align}
    \mathcal{L}(&D_s, D_c, \Psi_a, \Psi_g) = 
    \mathrm{E}_{z \sim p_z^g, \theta \sim p_\theta} [f(D_s(\Psi_g(z,\theta))]\nonumber\\
    &+\mathrm{E}_{ z_i \sim p_{z_i}^a, \theta \sim p_\theta} [f(D_s(\Psi_a(z_i,\theta))]\nonumber \\
    &+ \mathrm{E}_{I \sim p_{rim}} [f(-D_c(I)) + \lambda_{im} || \nabla D_c(I)||^2]\nonumber\\
    &+ \mathrm{E}_{L \sim p_{rsem}} [f(-D_s(L)) + \lambda_{s} || \nabla D_s(L)||^2]
    \label{eq:loss}
\end{align}

\textcolor{black}{ Here, $f(x) = log(1+exp(x))$ is softplus function, $I$ is a real image sample from $p_{rim}$ and corresponding annotated label $L \sim p_{rsem}$. Furthermore, to tackle concave geometry challenges and improve GAN inversion, we integrate two additional losses, namely, pose loss and latent loss outlined in Equations \ref{eq:pose} and \ref{eq:latent}, respectively. 
 These are combined with the loss presented in Equation \ref{eq:loss}.
 Additionally, it's worth noting that $\Psi_g$ learns to generate $L$ with $N$ labels, while $\Psi_a$ conditions clubbed semantics defined by $n$.
}
\begin{align}
    \mathcal{L}_p=& 
    \mathrm{E}_{ z_i \sim p_{z_i}^a, \theta \sim p_\theta}  ||f(D_c^p(\Psi_a(z_i,\theta)) - \theta||^2\nonumber\\
    &+\mathrm{E}_{I \sim p_{rim}} ||f(D_c^p(I) - \hat{\theta}||^2 
    \label{eq:pose}
\end{align}
\vspace{-0.75cm}
\begin{align}
    \mathcal{L}_{latent}=& 
    \mathrm{E}_{z \sim p_z^g, \theta \sim p_\theta} ||f(D_s^l(\Psi_g(z,\theta)) - z||^2
    \label{eq:latent}
\end{align}

\noindent \textcolor{black}{During first stage of training we incorporate all of the defined losses as given by Equation \ref{eq:s1loss}. }
\begin{align}
    \mathcal{L}_{S1}= \mathcal{L}(D_s, D_c, \Psi_a, \Psi_g) +   \lambda_{p}\mathcal{L}_p +  \lambda_{l}\mathcal{L}_{latent}
    \label{eq:s1loss}
\end{align}
\noindent Proceeding to second stage of training, we freeze weights of Geometry Module $\Psi_g$ and fine-tune the weights that affect appearance. The effective loss function in second stage is outlined in equation \ref{eq:s2loss}.
\begin{align}
    \mathcal{L}_{S2}= \mathcal{L}(D_c, \Psi_a) +   \lambda_{p}\mathcal{L}_p 
    \label{eq:s2loss}
\end{align}
\noindent where  $\mathcal{L}(D_c, \Psi_a)$ is given by:
\begin{align}
    \mathcal{L}(& D_c, \Psi_a) = 
    \mathrm{E}_{ z_i \sim p_{z_i}^a, \theta \sim p_\theta} [f(D_s(\Psi_a(z_i,\theta))]\nonumber \\
    &+ \mathrm{E}_{I \sim p_{rim}} [f(-D_c(I)) + \lambda_{im} || \nabla D_c(I)||^2]
    \label{eq:lossapp}
\end{align}
In our experiments, we empirically set the values of $\lambda_{im}=5, \lambda_s=1, \lambda_p=10$ and $\lambda_l= 1$, during first stage training. During second stage we use $\lambda_{im}=1$ and $\lambda_p= 10$.

\section{Pivotal Inversion} 
In this section we present mathematical formulation for optimizing $\delta\textit{w}^+$ and  $\delta \textit{w}_i^+$ to obtain geometric and appearance latent codes $w \in \mathcal{W}$ and $w_i \in \mathcal{W}_i$.
To edit an input image obtained either from $\mathrm{I} \sim p_{rim}$ (real images) or synthetically generated using Generators $\Psi_g$ and $\Psi_a$, we perform image inversion into the $\mathcal{W}$ space, represented as $\textit{w}$, utilizing pivotal tuning inversion \cite{roich2022pivotal}.
In the main paper, we described that to achieve editing; we optimize an editing offset vector $\delta\textit{w}^+ \in \mathcal{W}$ such that the generated semantic mask $S'$ from $\Psi_g$ approximates the original mask $S$. Similarly, for appearance, we optimize for $\delta \textit{w}_i^+$ to make the appearance resemble the given image $\mathrm{I}$.

To achieve the inversion of a real image $I$, we estimate the geometric and appearance offsets, $\delta\textit{w}^+$ and $\delta \textit{w}_i^+$ respectively, as articulated in Equation \ref{eq:ginv}.
\begin{align}
    \mathcal{L}(\delta\textit{w}^+, \delta \textit{w}_i^+)& = \lambda_s \mathcal{L}_s(S',S) \nonumber\\
    &+ \lambda_{im}\mathcal{L}_{im}(I',I) +  \lambda_{vgg}\mathcal{L}_{vgg}(I',I)
    \label{eq:ginv}
\end{align}
\noindent Here, $\mathcal{L}_s$ can be Cross-Entropy Loss or Mean squared Error Loss between predicted and ground truth semantics. $\mathcal{L}_{im}$ is Mean squared error loss and $\mathcal{L}_{vgg}$ is perceptual loss \cite{johnson2016perceptual} with VGG \cite{simonyan2014very} as backbone network for feature extraction. In our experiments, we setup $\mathcal{L}_s$ as Mean square error loss and $\lambda_s = 10, \lambda_{im}=1$, and $\lambda_{vgg}=1$.

\begin{figure}[ht]
\begin{center}
    \includegraphics[width=\linewidth]{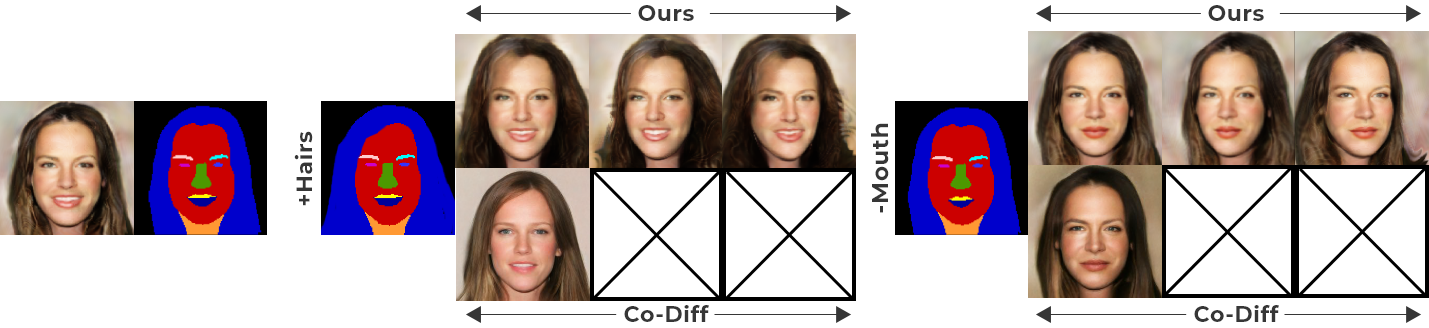}
\end{center}
    \vspace{-0.3cm}
    \caption{
    \textcolor{black}{Effect of manipulation in the semantic mask by expanding the hair region semantics and the effect of shrinking at the mouth region compared with Co-Diff \cite{huang2023collaborative}.
    }}
    \label{fig:edco}
    \vspace{-0.2cm}
\end{figure}

\subsection{Editing Faces and Facial Attribute Transfer}
When editing faces using a semantic mask or transferring semantic attributes represented by a mask $\mathcal{M}_k$, as detailed in the main paper, let $r$ denote the region requiring alteration while preserving the integrity of the remaining image portion. 
For editing generated image $I'$ given an edited semantic mask $S_{ed}$ corresponding to original semantics $S$. We modify inversion equation as presented in Equation \ref{eq:ginv}, following the formulation specified in Equation \ref{eq:edit}, to estimate $\delta\textit{w}^+$ and $\delta \textit{w}_i^+$ for the required edit. It's important to note that we only apply loss within the known regions, i.e., $(1-r)$, while granting the generator the flexibility to realistically fill the region $r$, thereby achieving the intended editing outcome.
In the context of transferring facial attributes between a given source image $I_s$ and a target image $I_t$, the inversion equation, articulated in Equation \ref{eq:ginv}, is similarly adjusted in line with the formulation presented in Equation \ref{eq:edit}. This adjustment serves the purpose of estimating $\delta\textit{w}^+$ and $\delta \textit{w}_i^+$, necessary for the attribute transfer.  
In our experiments, we setup $\mathcal{L}_s$ as Mean square error loss and $\lambda_s = 10, \lambda_{im}=1$, and $\lambda_{vgg}=1$. 
\begin{align}
    \mathcal{L}(\delta\textit{w}^+, \delta \textit{w}_i^+) &= \lambda_s \mathcal{L}_s(S',S_{ed}) \nonumber\\
    &+ \lambda_{im}\mathcal{L}_{im}(I'\odot (1-r),I\odot (1-r)) \nonumber\\
    &+ \lambda_{vgg}\mathcal{L}_{vgg}(I'\odot (1-r),I\odot (1-r))
    \label{eq:edit}
\end{align}
\begin{align}
    \mathcal{L}(\delta\textit{w}^+, \delta \textit{w}_i^+) &= \lambda_s \mathcal{L}_s(S' \odot (1-r),S_s \odot (1-r)) \nonumber\\
    &+ \lambda_s \mathcal{L}_s(S' \odot r,S_t \odot r) \nonumber\\
    &+ \lambda_{im}\mathcal{L}_{im}(I'\odot (1-r),I_s\odot (1-r)) \nonumber\\
    &+ \lambda_{vgg}\mathcal{L}_{vgg}(I'\odot (1-r),I_s\odot (1-r)) \nonumber\\
    &+ \lambda_{im}\mathcal{L}_{im}(I'\odot r,I_t\odot r) \nonumber\\
    &+ \lambda_{vgg}\mathcal{L}_{vgg}(I'\odot r,I_t \odot r)
    \label{eq:transfer}
\end{align}

\textcolor{black}{
\section{Qualitative Results}
In Figure \ref{fig:samples}, we show neural 72 rendering of generated semantic and rgb-radiances produced by latent codes randomly sampled from a gaussian distribution.
Furthermore, we provide a link to supplementary video containing GIFs showcasing multi-view renderings from different camera viewpoints: \url{https://youtube.com/shorts/b53EkPVK328?feature=share}
}

\begin{figure}[t]
\begin{center}
    \includegraphics[width=0.7\linewidth]{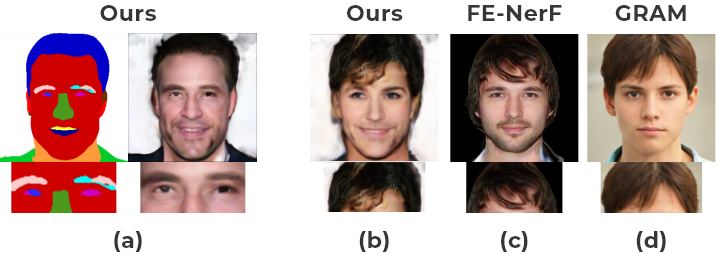}
\end{center}
    \vspace{-0.2cm}
    \caption{
    \textcolor{black}{Artifacts due to discrepancy in semantic radiance (a) and Hair quality obtained by different methods (c), (d) compared with Ours (b). 
    }}
    \label{fig:limits}
    \vspace{-0.3cm}
\end{figure}

\textcolor{black}{
\subsection{Semantics Guided Editing}
We present editing obtained by our method compared with Co-Diff \cite{huang2023collaborative} in Figure \ref{fig:edco}. Note that since our approach generates semantic and rgb-radiances in a volume, by design it supports generation of multi-view images by changing camera position. In contrast Co-Diff \cite{huang2023collaborative} employs a diffusion-based model to generate image in 2D space.
}

\textcolor{black}{
\subsection{Limitations}
The generation of RGB-radiance is influenced by the points grouped by the semantic volume masking layer. Consequently, any discrepancies in the generated semantic radiance propagate, leading to visual artifacts in the final image, as shown in Figure \ref{fig:limits}(a). Additionally, the generation of RGB-radiance is constrained to adhere to the densities of the semantic radiance, which limits the creation of fine-grained geometry, such as hairs.
We observe that the quality of hair generation in methods that learn the RGB-radiance field in conjunction with semantic radiances lags behind other 3D-aware GAN methods that do not incorporate semantics \cite{deng2022gram}.  For example, in Figure \ref{fig:limits}(b) and Figure \ref{fig:limits}(c), both our method and FE-NerF \cite{sun2022fenerf}, respectively, produce hair strands that are less realistic compared to those generated by Gram \cite{deng2022gram}, as illustrated in Figure \ref{fig:limits}(c).
}

\begin{figure}[ht]
\begin{center}
    \includegraphics[width=\linewidth]{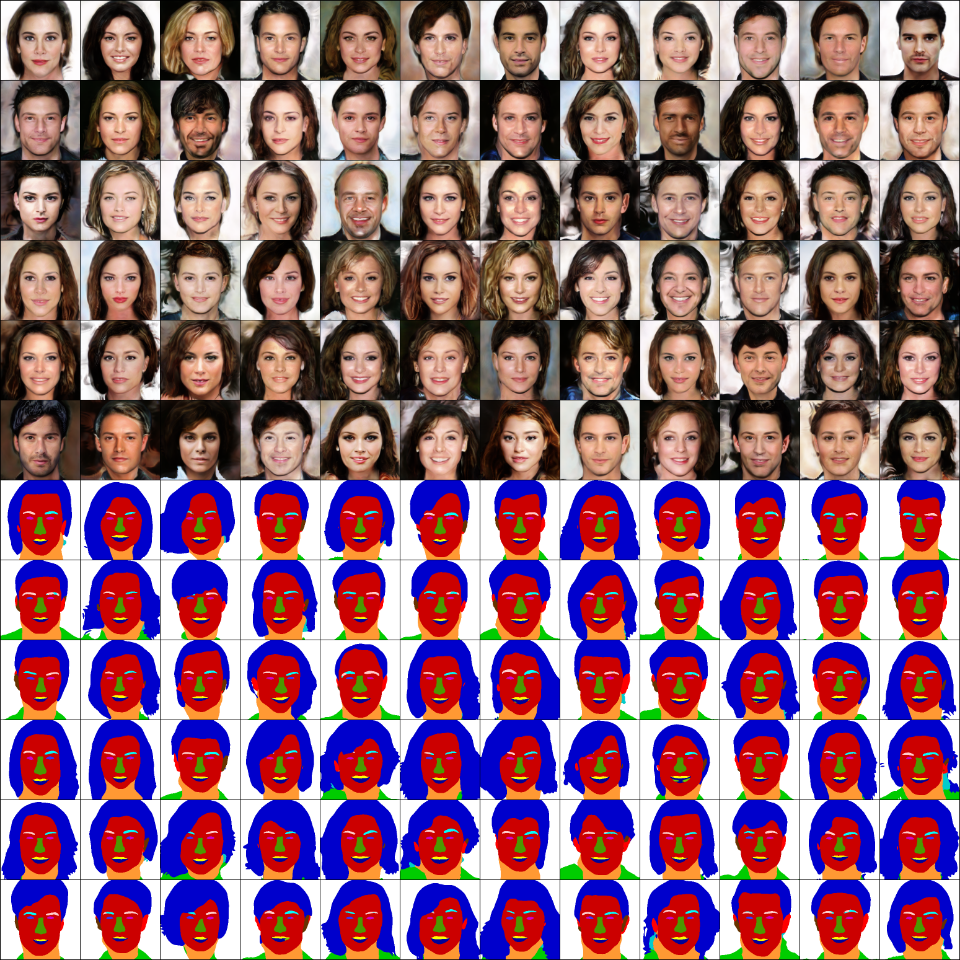}
\end{center}
    \caption{
    \textcolor{black}{Renderings of Semantic-radiance and RGB-radiance on image space generated by our approach by random latent code $\mathbf{z} \in \mathrm{R}^d$ and $\mathbf{z}_i \in  \mathrm{R}^d$. 
    }}
    \label{fig:samples}
    \vspace{-0.3cm}
\end{figure}


\textcolor{black}{
\section{Data Preprocessing}
We utilize the CelebAMask-HQ dataset \cite{CelebAMask-HQ} for training the proposed network because it includes semantic segmentation masks along with facial images. Initially, we crop and align all images from the dataset to center the facial region using the method described in \cite{bulat2017far}. We also adjust the semantic masks according to the alignment transformation of the corresponding images.
Furthermore, for each face, we estimate pose similar to \cite{deng2019accurate}. Alignment ensures that facial features (e.g., eyes, nose, mouth) are consistently positioned across all training images. This consistency along with camera pose information helps the model learn more accurate and generalized implicit representations of facial features and their spatial relationships in volumetric field.
}

\noindent \textcolor{black}{\textbf{Training on in the wild images.} To enhance diversity in terms of ethnicity and appearance, the model should be trained on a more varied dataset. When using images sourced from the web or repositories, the facial images must first be aligned as previously discussed. Subsequently, a semantic face parsing network, such as those described in \cite{CelebAMask-HQ}\cite{zheng2022general}\cite{yu2021bisenet} can be employed to create the semantic masks necessary for training.
}

\section{Ablation Studies}
\begin{table}[t]
\caption{
Comparing FID and KID between 5K generated and 5K real CelebA-MaskHQ \cite{CelebAMask-HQ} Dataset images with different weight sharing in Appearance Module. Each variant is trained for 60K iterations (2 epoch) jointly learning weights for both Geometry Module and Appearance Module (First stage).
}
\centering
\begin{tabular}{|l|r|r|}
\hline
\multicolumn{1}{|c|}{\textbf{Method}} & \multicolumn{1}{c|}{\textbf{FID} $\downarrow$} & \multicolumn{1}{c|}{\textbf{KID} $(\times 10^3)$ $\downarrow$} \\ \hline
No shared weights & 30.36 & 41.62 \\ \hline
\begin{tabular}[c]{@{}l@{}}Fully Shared AM \\ (shared Linear Color layers)\end{tabular} & 24.43 & 36.32 \\ \hline
\begin{tabular}[c]{@{}l@{}}Proposed Sharing \\ (SemFaceEdit)\end{tabular}  & 22.65 & 34.74 \\ \hline
\end{tabular}
\label{tab:ablshare}
\end{table}

\begin{table}[t]
\caption{Comparing FID and KID between 5K generated and 5K real CelebA-MaskHQ \cite{CelebAMask-HQ} Dataset images with different architecture depth of Geometry Module and Appearance Module. Each variant is trained for 60K iterations (2 epochs), jointly learning weights for both Geometry Module and Appearance Module (First stage).}
\centering
\begin{tabular}{|cccc|}
\hline
\multicolumn{4}{|c|}{\textbf{Depth}} \\ \hline
\multicolumn{1}{|c|}{\textbf{Geometric Module}} & \multicolumn{1}{c|}{\textbf{Appearance Module}} & \multicolumn{1}{c|}{\textbf{FID $\downarrow$}} & \textbf{KID $(\times 10^3)$ $\downarrow$} \\ \hline
\multicolumn{1}{|c|}{4} & \multicolumn{1}{c|}{4} & \multicolumn{1}{c|}{27.18} & 38.35 \\ \hline
\multicolumn{1}{|c|}{6} & \multicolumn{1}{c|}{6} & \multicolumn{1}{c|}{24.12} & 35.43 \\ \hline
\multicolumn{1}{|c|}{8} & \multicolumn{1}{c|}{8} & \multicolumn{1}{c|}{22.65} & 34.74 \\ \hline
\end{tabular}%
\label{tab:abldepth}
\end{table}

In the proposed architecture \textit{(Figure $3$ of main paper)}, we implement shared weights for the Appearance Module (excluding Linear Color Layers) across all semantic categories. This design choice not only reduces model complexity but also contributes to enhanced metrics, as demonstrated in Table \ref{tab:ablshare}. Further, we perform experiments by varying the number of FiLM Layers \cite{chan2021pi} in both the Geometry and Appearance Module.
It's noteworthy that we observe improvements in performance corresponding to an increase in the number of layers within the neural networks, as illustrated in Table \ref{tab:abldepth}. In consideration of computational constraints and model complexity, the results presented from SemFaceEdit are based on a depth value of $8$.

\end{document}